\documentclass[runningheads]{llncs}
\usepackage[T1]{fontenc}
\usepackage{times}
\usepackage{bm}
\usepackage{tabularx}
\usepackage{epsfig}
\usepackage{calc}
\usepackage{amssymb}
\usepackage{amstext}
\usepackage{siunitx}
\usepackage{booktabs}
\usepackage{algorithmic}
\usepackage[table,xcdraw]{xcolor}
\usepackage{amsmath}
\usepackage[export]{adjustbox}
\usepackage{multicol}
\usepackage{multirow}
\usepackage{textcomp}
\usepackage{pslatex}
\usepackage{float}
\usepackage[normalem]{ulem}
\useunder{\uline}{\ul}{}

\usepackage{arydshln}
\usepackage[space]{cite}
\usepackage[normalem]{ulem}
\usepackage[symbol]{footmisc}
\newcommand{\minus}{\scalebox{0.5}[1.0]{$-$}}
\usepackage[pagebackref=true,breaklinks=true,colorlinks,bookmarks=false]{hyperref}
\def\BibTeX{{\rm B\kern-.05em{\sc i\kern-.025em b}\kern-.08em
    T\kern-.1667em\lower.7ex\hbox{E}\kern-.125emX}}

\begin{document}

\title{Transformers in Unsupervised Structure-from-Motion}

\titlerunning{MT-SfMLearner v2}

\author{Hemang Chawla$^{*1,2}$\orcidID{0000-0002-5999-6901} \and
Arnav Varma$^{*1}$\orcidID{0000-0002-5919-0449} \and \\
Elahe Arani$^{1,2}$\orcidID{0000-0002-0952-7007} \and
Bahram Zonooz$^{1,2}$\orcidID{0000-0003-4124-3394}}

\authorrunning{H. Chawla et al.}

\institute{
$^{1}$Advanced Research Lab, NavInfo Europe, Eindhoven, The Netherlands \\
$^{2}$Department of Mathematics and Computer Science, Eindhoven University of Technology, Eindhoven, The Netherlands \\
\email{\{firstname.lastname\}@navinfo.eu, bahram.zonooz@gmail.com} \\
\url{https://www.navinfo.eu/expertise/artificial-intelligence/}
}

\maketitle
\def\thefootnote{*}\footnotetext{Equal contribution.}\def\thefootnote{footnote}

\begin{abstract}
\renewcommand{\thefootnote}{\fnsymbol{footnote}}
Transformers have revolutionized deep learning based computer vision with improved performance as well as robustness to natural corruptions and adversarial attacks. Transformers are used predominantly for 2D vision tasks, including image classification, semantic segmentation, and object detection. However, robots and advanced driver assistance systems also require 3D scene understanding for decision making by extracting structure-from-motion (SfM).
We propose a robust transformer-based monocular SfM method that learns to predict monocular pixel-wise depth, ego vehicle's translation and rotation, as well as camera's focal length and principal point, simultaneously. With experiments on KITTI and DDAD datasets, we demonstrate how to adapt different vision transformers and compare them against contemporary CNN-based methods. 
Our study shows that transformer-based architecture, though lower in run-time efficiency, achieves comparable performance while being more robust against natural corruptions, as well as untargeted and targeted attacks.\footnote[4]{Code: \url{https://github.com/NeurAI-Lab/MT-SfMLearner}}

\end{abstract}

\section{Introduction}
Scene understanding tasks have benefited immensely from advances in deep learning over the years~\cite{yao2012describing}. Several existing methods in computer vision for robotics~\cite{li2022towards}, augmented reality~\cite{kastner20203d}, and autonomous driving~\cite{peng2020imitative} have been using convolutional neural networks (CNNs) with its properties of spatial locality and translation invariance~\cite{lecun1998gradient} resulting in excellent performance. With the advent of vision transformers~\cite{vaswani2017attention, dosovitskiy2020vit}, models with the ability to learn from the global context have even outperformed CNNs for some tasks such as object detection~\cite{carion2020end} and semantic segmentation~\cite{zheng2021rethinking}. Nevertheless, despite performance improvements in curated test sets, safe deployment of models requires further consideration of robustness and generalizability. 

Different neural network architectures have been shown to have different effects on model performance, robustness, and generalizability on different tasks~\cite{huang2021exploring, croce2022interplay}. CNNs have localized linear operations and lose feature resolution during downsampling to increase their limited receptive field~\cite{yang2021transformers}. Transformers, with their different layers that simultaneously attend more to global features with no inductive bias in favor of locality lead to more globally coherent predictions~\cite{touvron2021training}.
Among these architectures, transformers have been found to be more robust for tasks such as classification~\cite{bhojanapalli2021understanding, raghu2021vision}, object detection, and semantic segmentation~\cite{jeeveswaran2022comprehensive} despite requiring more training data and being more computationally expensive~\cite{caron2021emerging}. 
Although several studies have compared their performance on 2D vision tasks~\cite{jeeveswaran2022comprehensive}, studies that evaluate their performance on 3D scene understanding tasks such as monocular Structure-from-Motion (SfM) are lacking.

SfM is a prominent problem in 3D computer vision in which the 3D structure is reconstructed by simultaneously estimating scene depth, camera poses, and intrinsic parameters. Traditional methods for SfM rely on correspondence search followed by incremental reconstruction of the environment, and are able to handle a diverse set of scenes~\cite{schonberger2016structure}. Instead, most deep learning based methods for SfM primarily focus on depth estimation with pose and intrinsics estimation as auxiliary tasks~\cite{guizilini20203d, lyu2020hr}. Deep learning methods have increasingly been employed, either in replacement or together with traditional methods, to deal with issues of low-texture, thin structures, misregistration, failed image registration, etc.~\cite{zhou2017unsupervised}. While most of the existing works for depth estimation were based on CNNs~\cite{godard2017unsupervised, ming2021deep, guizilini20203d, lyu2020hr}, transformer-based approaches to supervised depth estimation have also been proposed~\cite{Ranftl2020}. However, unsupervised methods that do not require ground truth collected from costly LiDARs or RGB-D setups are often favored, as they can potentially be applied to innumerable data. Methods that utilize transformer ingredients~\cite{johnston2020self} such as attention have been proposed to improve depth estimation, but transformer encoders have scarcely been adopted for depth and pose estimation. 

In this work, we perform a comparative analysis between CNN- and transformer-based architectures for unsupervised depth estimation. We show how vision transformers can be adapted for \textit{unsupervised} depth estimation with our method Monocular Transformer SfMLearner (MT-SfMLearner). We evaluate how the architecture choices for individual depth and pose estimation networks impact the depth estimation performance, as well as robustness to natural corruptions and adversarial attacks. Since SfM depends upon the knowledge of camera intrinsics, we also introduce a modular approach to predict the camera focal lengths and principal point from the input images, which can be utilized within both CNN- and transformer-based architectures. We also study the accuracy of intrinsics and pose estimation, including the impact of learning camera intrinsics on depth and pose estimation. Finally, we compare the computational and energy efficiency of the architectures for depth, pose, and intrinsics estimation.

This work is an extended version of our study comparing transformers and CNNs for unsupervised depth estimation~\cite{visapp22}. While the previous work demonstrated how vision transformers that were built for classification-like tasks 
can be also used for unsupervised depth estimation, here we demonstrate that our method also extends to other transformer architectures. 
We also perform an additional evaluation on a more challenging dataset to further substantiate the generalizability of our study. Additionally, we also compare the impact of architectures on the performance of the auxiliary pose prediction task, including when the camera instrincs are learned simultaneously. With a more general purpose method, additional experiments, quantitative results, and visualizations, this work presents a way to compare the trade-off between the performance, robustness, and efficiency of transformer- and CNN-based architectures for monocular unsupervised depth estimation.

\section{Related Works}

The simultaneous estimation of Structure-from-Motion (SfM) is a well-studied problem with an established toolchain of techniques~\cite{schonberger2016structure}. Although the traditional toolchain is effective and efficient in many cases, its reliance on accurate image correspondence can cause problems in areas of low texture, complex geometry/photometry, thin structures, and occlusions~\cite{zhou2017unsupervised}. To address these issues, several of the pipeline stages have recently been tackled using deep learning, e.g., feature matching~\cite{jiang2022glmnet}, pose estimation~\cite{bian2021unsupervised}, and stereo and monocular depth estimation~\cite{godard2017unsupervised}. Of these, unsupervised monocular depth estimation in particular has been extensively explored with CNN-based depth and pose networks, with pose estimation as an auxiliary task~\cite{zhou2017unsupervised, godard2017unsupervised, lyu2020hr, guizilini20203d, bae2022monoformer}. These learning-based techniques are attractive because they can utilize external supervision during training and may circumvent the aforementioned issues when applied to test data. However, learned systems might not be robust to shifts in distribution during test time.

Recently, transformer-based architectures~\cite{dosovitskiy2020vit, touvron2021training}, which outperform CNN-based architectures in image classification, have been proven to be more robust in image classification~\cite{bhojanapalli2021understanding,paul2021vision}, as well as dense prediction tasks such as object detection and semantic segmentation~\cite{jeeveswaran2022comprehensive}. Motivated by their success, researchers have replaced CNN encoders with transformers in scene understanding tasks such as object detection~\cite{carion2020end,liu2021swin}, semantic segmentation~\cite{zheng2021rethinking,strudel2021segmenter}, and supervised monocular depth estimation~\cite{Ranftl2020,yang2021transformers}. Our previous work~\cite{visapp22} and MonoFormer~\cite{bae2022monoformer} further extend transformer-based architectures to unsupervised monocular unsupervised depth estimation. However, ours is the only work that comprehensively demonstrates the robustness of transformer-based architectures for unsupervised SfM. 
We now provide analyses to establish generalizability of our approach across multiple datasets, transformer-based architectures, and auxiliary SfM tasks such as pose estimation and intrinsics estimation.

\section{Method}
\label{sec:method}
We study the impact of using vision transformer based architectures for unsupervised monocular Structure-from-Motion (SfM) in contrast to contemporary methods that utilize CNN-based architectures.

\subsection{Monocular Unsupervised SfM}
\label{subsec:mono_sfm}

\begin{figure*}[t]
\centering

  \includegraphics[width=\textwidth]{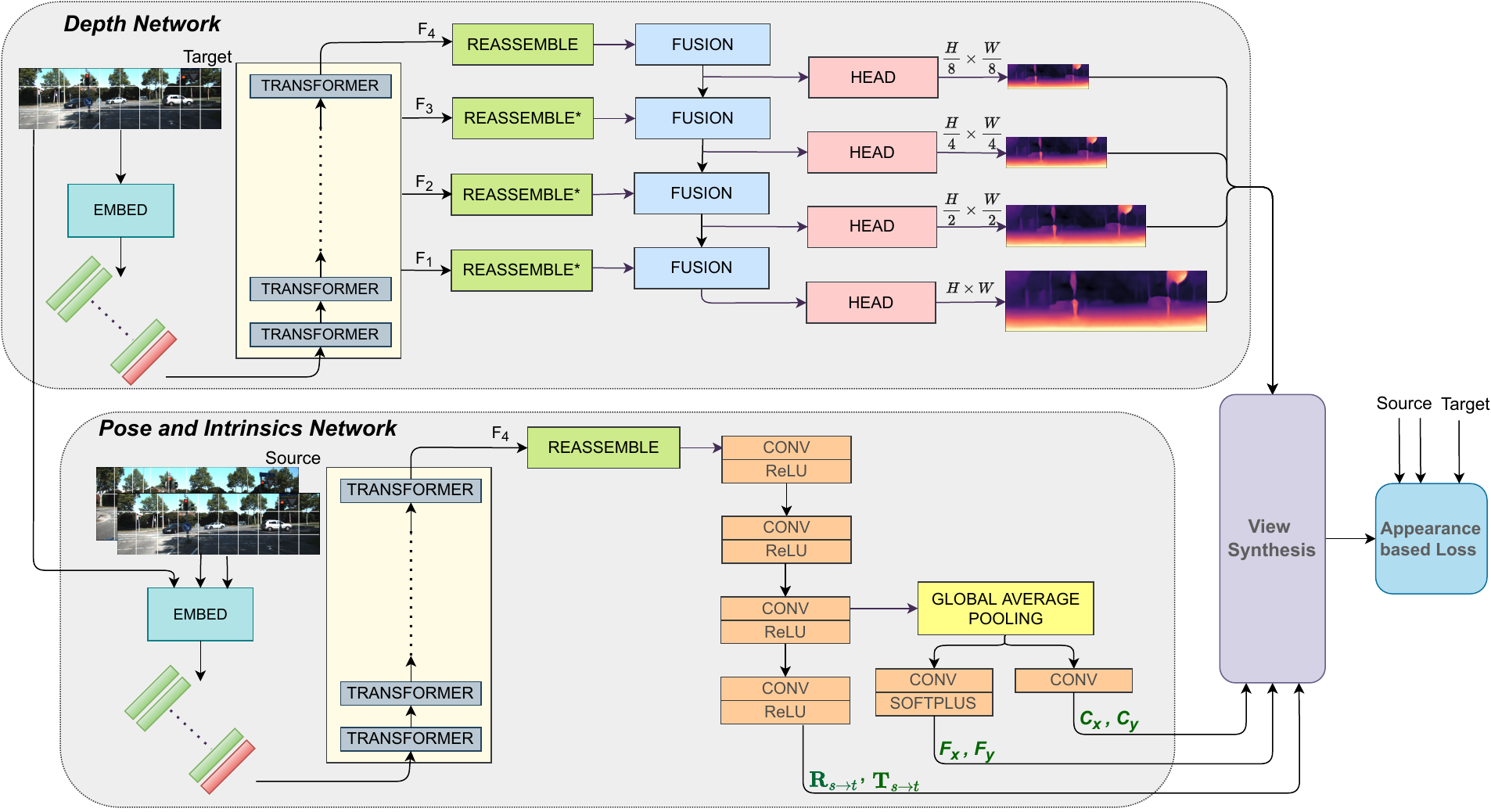}
  \caption{An overview of Monocular Transformer Structure from Motion Learner (MT-SfMLearner) with learned intrinsics. We readapt modules from  Dense Prediction Transformer (DPT) and Monodepth2 to be trained with appearance-based losses for unsupervised monocular depth, pose, and intrinsics estimation. \textbf{*} refers to optional modules that may not be present in all transformer-based architectures. Figure modified from ~\cite{visapp22}.} 
\label{fig:architecture}
\end{figure*}

For training unsupervised SfM networks, we utilize videos captured from monocular cameras. Given a video sequence with $n$ images, both the depth and pose estimation networks are trained simultaneously. This is unlike supervised networks, where depth or pose estimation may be trained independently. The input to the depth estimation network $f_D:\mathbb{R}^{H\times W \times 3}$ is a single image, for which it outputs pixel-wise depth in $\mathbb{R}^{H \times W}$. The input to the pose estimation network $f_{E}:\mathbb{R}^{H\times W \times 6}$ is a pair of images, for which it outputs the relative translation $(t_x, t_y, t_z)$  and rotation $(r_x, r_y, r_z) \in \mathbb{R}^6$, which is used to form the affine transformation \tiny $\begin{bmatrix} \hat{{R}} & \hat{{T}} \\ {0} & 1 \end{bmatrix}$ \normalsize $\in \text{SE(3)}$ . To train both networks simultaneously, a batch consists of triplets of temporally consecutive RGB images $\{I_{\minus 1}, I_0, I_1\} \in \mathbb{R}^{H\times W \times 3}$. While $I_0$ is input into the depth estimation network, $\{I_{\minus 1}, I_0\}$ and $\{I_0, I_1\}$ are input into the pose estimation network to predict the next and previous relative pose. The perspective projection model links together the predicted depth $\hat{{D}}$ and pose $\hat{{T}}$ such that,  
\begin{equation}
\label{eq:perspective_model}
    p_s \sim K\hat{{R}}_{s \leftarrow t} \hat{{D}}_t(p_t)K^{-1}p_t + K\hat{{T}}_{s\leftarrow t}. 
\end{equation}
This is used to warp the source images $I_s \in \{I_{\minus 1}, I_1\}$ to the target image $I_t \in \{I_0\}$ as part of the view synthesis (see Figure~\ref{fig:architecture}), where $K$ represents the camera intrinsics. For each triplet, two target images $\hat{I}_{0}$ are synthesized, which are compared with the real target image, to compute the appearance-based \textit{photometric} loss. Additionally, we utilize a smoothness loss~\cite{guizilini20203d} on the predicted depth for regularization.

\subsection{Architecture}
\label{subsec:arch}

\begin{table}[t]
\centering
\caption{Architecture details of the \textit{Reassemble} modules. $DN$ and $PN$ refer to depth and pose networks, respectively. The subscripts of $DN$ refer to the transformer stage from which the respective \textit{Reassemble} module takes its input (see Figure \ref{fig:architecture}). The input image size is $H \times W$, $p$ refers to the patch size, $N_p=H\cdot W / p^2$ refers to the number of patches in the image, $s$ refers to the stride of the \textit{Embed} module and $d$ refers to the feature dimension of the transformer features. Table modified from~\cite{visapp22}.}
\label{tab:reassemble}
\begin{tabular}{|l|l|l|l|l|l|}
\hline
\begin{tabular}[c]{@{}l@{}}\textbf{Encoder}\end{tabular} & \textbf{Operation} & \textbf{Input size} & \textbf{Output size} & \textbf{Function} & \begin{tabular}[c]{@{}l@{}}\textbf{Parameters}\\ ($DN_1$, $DN_2$, \\$DN_3$, $DN_4$, \\$PN_4$)\end{tabular} \\ \hline \hline
\multirow{15}{*}{\textbf{DeiT}}                                                    
 & Read  &    $(N_p + 1) \times d$   &   $N_p \times d$  &  Drop readout token   &   $-$ \\ \cline{2-6}

 & Concatenate   &    $N_p \times d$  &    $d \times \frac{H}{p} \times \frac{W}{p}$  &   \begin{tabular}[c]{@{}l@{}}Transpose \& Unflatten\end{tabular} & $-$ \\ \cline{2-6}

 & \begin{tabular}[c]{@{}l@{}}Pointwise\\ Convolution\end{tabular} &  $d \times \frac{H}{p} \times \frac{W}{p}$  &  $N_c \times \frac{H}{p} \times \frac{W}{p}$  &  Change to $N_c$ channels  & 
 \begin{tabular}[c]{@{}l@{}}$N_c=$\\$[96, 768,$ \\$1536, 3072,$ \\$2048]$\end{tabular} \\ \cline{2-6}

 & \begin{tabular}[c]{@{}l@{}}Strided\\ Convolution\end{tabular}   &  $N_c \times \frac{H}{p} \times \frac{W}{p}$  & $N_c \times \frac{H}{2p} \times \frac{W}{2p}$  
 & \begin{tabular}[c]{@{}l@{}}$k \times k$ convolution, stride$=2$, \\ 
 $N_c$ channels, padding$=1$\end{tabular} & 
 \begin{tabular}[c]{@{}l@{}}$k=$\\$[-, -,$\\ $-, 3,$\\ $-]$\end{tabular} \\ \cline{2-6}

 & \begin{tabular}[c]{@{}l@{}}Transpose\\ Convolution\end{tabular} & $N_c \times \frac{H}{p} \times \frac{W}{p}$  &  $N_c \times \frac{H}{\alpha} \times \frac{W}{\alpha}$ 
 & \begin{tabular}[c]{@{}l@{}} $\frac{p}{\alpha} \times \frac{p}{\alpha}$ deconvolution, \\ 
 stride$=\frac{p}{\alpha}$, $N_c$ channels\end{tabular}  & 
 \begin{tabular}[c]{@{}l@{}}$\alpha=$\\$[4, 8,$\\ $-, -,$\\ $-]$\end{tabular} \\ \hline \hline
\textbf{PVT} & Reshape  & $\frac{HW}{64s^2} \times N_b$  & $N_b \times \frac{H}{8s} \times \frac{W}{8s}$ 
& \begin{tabular}[c]{@{}l@{}}Reshape token to image-like \\ $2D$ representations. \end{tabular}
& \begin{tabular}[c]{@{}l@{}}$N_b=$\\$[-, -,$\\ $-, 512,$\\ $512]$\end{tabular}\\ \hline
\end{tabular}
\end{table}

\subsubsection{Depth Network}
\label{subsubsec:depthnet}
For the depth network, we use a transformer-based architecture in the encoder, and readapt the decoder from the DPT~\cite{Ranftl2021}.
There are five components of the depth network:
\begin{itemize}
    \item \textbf{\textit{Embed}} module, which is part of the encoder, takes an image $I \in \mathbb{R}^{H\times W \times 3}$ and converts image patches of size $p \times p$ to $N_{p} = H \cdot W / p^2$ tokens $t_i \in \mathbb{R}^{d} \ \forall i \in [1, 2,... N_p]$. This is implemented as a $p \times p$ convolution with stride $s \leq p$. The output of this module may be concatenated with a \textit{readout} token $\in \mathbb{R}^d$, depending on the transformer based architecture.
    \item \textbf{\textit{Transformer}} block, which is also part of the encoder, consists of multiple transformer stages that process these tokens with self-attention modules~\cite{vaswani2017attention}. Self-attention processes inputs at constant resolution and can simultaneously attend to global and local features.
    \item \textbf{\textit{Reassemble}} modules in the decoder are responsible for extracting image-like $2D$ representations from the features of the transformer block. At least one reassemble module is used, and additional modules may be used depending on the transformer-based architecture. The exact details of the \textit{Reassemble} modules can be found in Table \ref{tab:reassemble}. 
    \item \textbf{\textit{Fusion}} modules in the decoder, based on RefineNet~\cite{lin2017refinenet}, are responsible for progressively fusing the features of the encoder or the \textit{Reassemble} modules with the features of the decoder. The module is also responsible for upsampling the features by $2$ at each stage. Unlike DPT, we enable batch normalization in the decoder, as it was found to be helpful for unsupervised depth prediction. We also reduce the number of channels in the \textit{Fusion} block to $96$ from $256$ in DPT. 
    \item \textbf{\textit{Head}} modules after each \textit{Fusion} module predict depth on four scales, according to previous unsupervised methods \cite{godard2019digging}. Unlike DPT, \textit{Head} modules use $2$ convolutions instead of $3$ as we did not find any difference in performance. For the exact architecture of the \textit{Head} modules, see Table \ref{tab:heads}.
\end{itemize}

\begin{table}[t]
    \centering
    \caption{Architecture details of \textit{Head} modules in Figure \ref{fig:architecture}. Source:~\cite{visapp22}.}
\begin{tabular}{|c|}
\hline
\textbf{Layers}                                                                            \\ \hline \hline
 $32$ $3 \times 3$ \textit{Convolutions}, stride=$1$, padding$=1$ \\ \hdashline
\textit{ReLU}                                                    \\ \hdashline
\textit{Bilinear Interpolation} to upsample by $2$                \\ \hdashline
$32$ \textit{Pointwise Convolutions}                                                   \\ \hdashline
\textit{Sigmoid}                                                 \\ \hline
\end{tabular}
    \label{tab:heads}
\end{table}

\subsubsection{Pose Network.}
\label{subsubsec:posenet}
For the pose network, we adopt an architecture similar to that of the depth network, with a transformer-based architecture in the encoder, but the decoder from Monodepth2~\cite{godard2019digging}. Since the input
to the transformer for the pose network consists of two images concatenated along the channel dimension, we repeat the \textit{Embed} module accordingly. Unlike the depth network, we only use a single \textit{Reassemble} module to pass transformer tokens to the decoder, independently of the transformer-based architecture used. For details of the structure of this Reassemble module, refer to Table~\ref{tab:reassemble}.

When both depth and pose networks use transformers as described above, we refer to the resulting architecture as Monocular Transformer Structure-from-Motion Learner (\textit{MT-SfMLearner}).

\subsection{Intrinsics}
\label{subsec:intrinsics}
As seen in Equation~\ref{eq:perspective_model}, unsupervised monocular SfM requires knowledge of the ground truth camera intrinsics. Intrinsics are given by
\begin{equation}
\label{eq:camera_matrix}
    K = \begin{bmatrix}
    f_x & 0 & c_x\\
    0 & f_y & c_y\\
    0 & 0 & 1
    \end{bmatrix},
\end{equation}
where $f_x$ and $f_y$ refer to the focal lengths of the camera along the $x$-axis and $y$-axis, respectively. $c_x$ and $c_y$ refer to $x$ and $y$ coordinates of the principal point in the pinhole camera model.
Most unsupervised SfM methods are trained with intrinsics known a priori. However, the intrinsics may vary within a dataset with videos collected from different camera setups or over a long period of time. These parameters can also be unknown for crowdsourced datasets.  

Therefore, we introduce an intrinsics estimation module. We modify the pose network to additionally estimate the focal lengths and principal point along with the translation and rotation.
Concretely, we add a convolutional path in the pose decoder to learn the intrinsics. The features before activation from the penultimate decoder layer are passed through a global average pooling layer. This is followed by two branches of pointwise convolutions that reduce the number of channels from $256$ to $2$. One branch uses softplus activation to estimate focal lengths along the $x$ and $y$ axes, as the focal length is always positive. The other branch estimates the principal point without employing any activation, as the principal point does not have such a constraint. Note that the pose decoder is the same for both CNN- and transformer-based architectures. Consequently, the intrinsics estimation method can be modularly utilized with both architectures. Figure \ref{fig:architecture} demonstrates MT-SfMLearner with learned intrinsics.

\subsection{Appearance-based Losses}
\label{subsec:appearance}
Following contemporary unsupervised monocular SfM methods, we adopt the \textit{appearance-based losses} and an \textit{auto-masking} procedure from CNN-based Monodepth2~\cite{godard2019digging} for the transformer-based architecture described above. 
We employ a photometric per-pixel minimum reprojection loss composed of the pixel-wise $\ell_1$ distance as well as the Structural Similarity (SSIM) between the real and synthesized target images, along with a multiscale edge-aware \textit{smoothness} loss on the depth predictions. We also use auto-masking to disregard the temporally stationary pixels in the image triplets. Finally, to reduce texture-copy artifacts, we calculate the total loss after upsampling the depth maps, predicted at 4 scales by the decoder, to the input resolution.
\section{Experiments}
\label{sec:exp}
We compare the CNN and transformer architectures for their impact on unsupervised monocular depth and pose estimation, including when the camera intrinsics are unknown and when they are estimated simultaneously. 

\subsection{Datasets}
\label{subsec:datasets}

\subsubsection{KITTI.} 
For depth estimation, we report results on the Eigen Split~\cite{eigen2014depth} of KITTI~\cite{geiger2013vision} dataset after removing the static frames as per \cite{zhou2017unsupervised}, unless stated otherwise. This split contains $39,810$ training images, $4424$ validation images, and $697$ test images, respectively. This dataset captures scenes from rural, city and highway areas around Karlsruhe, Germany. All results are reported on the per-image scaled dense depth prediction without post-processing~\cite{godard2019digging} for an image size of 640$\times$192, unless otherwise stated. 

For pose estimation in Section~\ref{subsec:generalizability}, we report results on the Odom Split~\cite{zhou2017unsupervised} of the KITTI dataset for an image size of 640$\times$192, unless stated otherwise. This split contains $8$ training sequences (sequences $00-02$, $04-08$) and two test sequences $(09, 10)$.

\subsubsection{Dense Depth for Autonomous Driving.}
For depth estimation, we also report results on the Dense Depth for Autonomous Driving (DDAD) dataset~\cite{guizilini20203d}, for an image size of 640$\times$ 384 unless otherwise noted. It contains $12650$ training samples from $150$ sequences and $3950$ test samples from $50$ sequences, respectively. This dataset contains samples with long range depth (up to 250 m) from a diverse set of urban scenarios in multiple cities of the United States (Ann Arbor, Bay Area, Cambridge, Detroit, San Francisco) and Japan (Odaiba, Tokyo). 

\subsection{Architecture}
\label{subsec:arch_expt}

For the transformer-based architecture in our depth and pose encoders, we use DeiT-base~\cite{touvron2021training} except in Section~\ref{subsec:generalizability}, where we use PVT-b4~\cite{wang2022pvt} to demonstrate that our approach generalizes to other transformer-based architectures. The \textit{Embed} module of DeiT-base has a patch size $p=16$ and stride $s=16$, while that of PVT-b4 has a patch size $p=7$ and a stride $s=4$. DeiT-base employs $12$ \textit{Transformer} stages with features $F_1$, $F_2$, $F_3$, and $F_4$ (see Figure~\ref{fig:architecture}) taken from the $3^{rd}$, $6^{th}$, $9^{th}$, and final stages to be sent to the decoder. PVT-b4, meanwhile, employs $4$ transformer stages, each of which contributes to the features sent to the decoder. Finally, DeiT-base uses $4$ \textit{Reassemble} modules in the depth encoder, while PVT-b4 uses only one \textit{Reassemble module} in the depth encoder. The exact architecture of the Reassemble modules can be found in Table~\ref{tab:reassemble}.

\subsection{Implementation Details}
\label{subsec:details}
The networks are implemented in PyTorch~\cite{paszke2019pytorch} and trained on a TeslaV$100$ GPU for $20$ epochs at a resolution of $640\times192$ with batch sizes $12$ for DeiT-base encoder and $8$ for PVT-b4 encoder, unless otherwise mentioned. The depth and pose encoders are initialized with ImageNet~\cite{deng2009imagenet} pre-trained weights. We use the Adam~\cite{kingma2014adam} optimizer for CNN-based networks and AdamW~\cite{loshchilov2017decoupled} optimizer for transformer-based networks with initial learning rates of $1e^{-4}$ and $1e^{-5}$, respectively. 
The learning rate is decayed after $15$ epochs by a factor of $10$. Both optimizers use $\beta_{1}=0.9$ and $\beta_{2}=0.999$.

\subsection{Evaluation Metrics}
\label{subsec:metrics}

\paragraph{Depth Estimation.} We measure the error and accuracy of depth estimation using various metrics. For error, we use the absolute relative error (Abs Rel)~\cite{saxena2008make3d}, squared relative error~\cite{koch2018evaluation} (Sq Rel), linear root mean squared error (RMSE)~\cite{li2010towards}, log scale invariant RMSE~\cite{eigen2014depth} (RMSE log). For accuracy, we measure under three thresholds, reported as ratios~\cite{ladicky2014pulling} ($\delta < 1.25$, $\delta < 1.25^2$, $\delta < 1.25^3$). 

\paragraph{Pose Estimation.} We measure translation and rotational errors for all possible subsequences of length $(100, 200, \dots, 800)m$, and report the average of the values. The translation error is reported as a percentage and the rotation error as degrees per $100m$~\cite{Geiger2012CVPR}.

\paragraph{Intrinsics Estimation.} We measure the percentage error from its ground truth value for each camera intrinsic parameter. 

\paragraph{Efficiency.} We measure computational and energy efficiency using frames per second (fps) and Joules per frame, respectively. 

\subsection{Impact of Architecture}

\begin{table}[ht]
\centering
\caption{Quantitative results on KITTI Eigen split for all four architecture combinations of depth and pose networks. The best results are displayed in bold, and the second best are underlined. Source:~\cite{visapp22}.} 
\begin{tabular}{|c|c|c|c|c|c|c|c|}
\hline
\multirow{2}{*}{\textbf{Architecture}}  & \multicolumn{4}{c|}{\textbf{\cellcolor{red!25}Error$\downarrow$}} & \multicolumn{3}{c|}{\textbf{\cellcolor{blue!25}Accuracy$\uparrow$}} \\ \cline{2-8}
       & Abs Rel & Sq Rel & RMSE & RMSE log & $\delta<1.25$ & $\delta<1.25^2$ & $\delta<1.25^3$ \\ \hline \hline
  C, C & \textbf{0.111} & 0.897 & 4.865 & 0.193 & \textbf{0.881} & \underline{0.959} & 0.980 \\ \hdashline
  C, T & 0.113 & 0.874 & 4.813 & 0.192 & \underline{0.880} & \textbf{0.960} & \underline{0.981} \\
  T, C  & \underline{0.112} & \underline{0.843} & \textbf{4.766} & \underline{0.189} & 0.879 & \textbf{0.960} & \textbf{0.982} \\ \hdashline
  T, T  & 0.112 & \textbf{0.838} & \underline{4.771} & \textbf{0.188} & 0.879 & \textbf{0.960} & \textbf{0.982} \\
  \hline
\end{tabular}
\label{tab:ablation_table}
\end{table}

\begin{figure}[ht]
\centering
    \includegraphics[width=\linewidth]{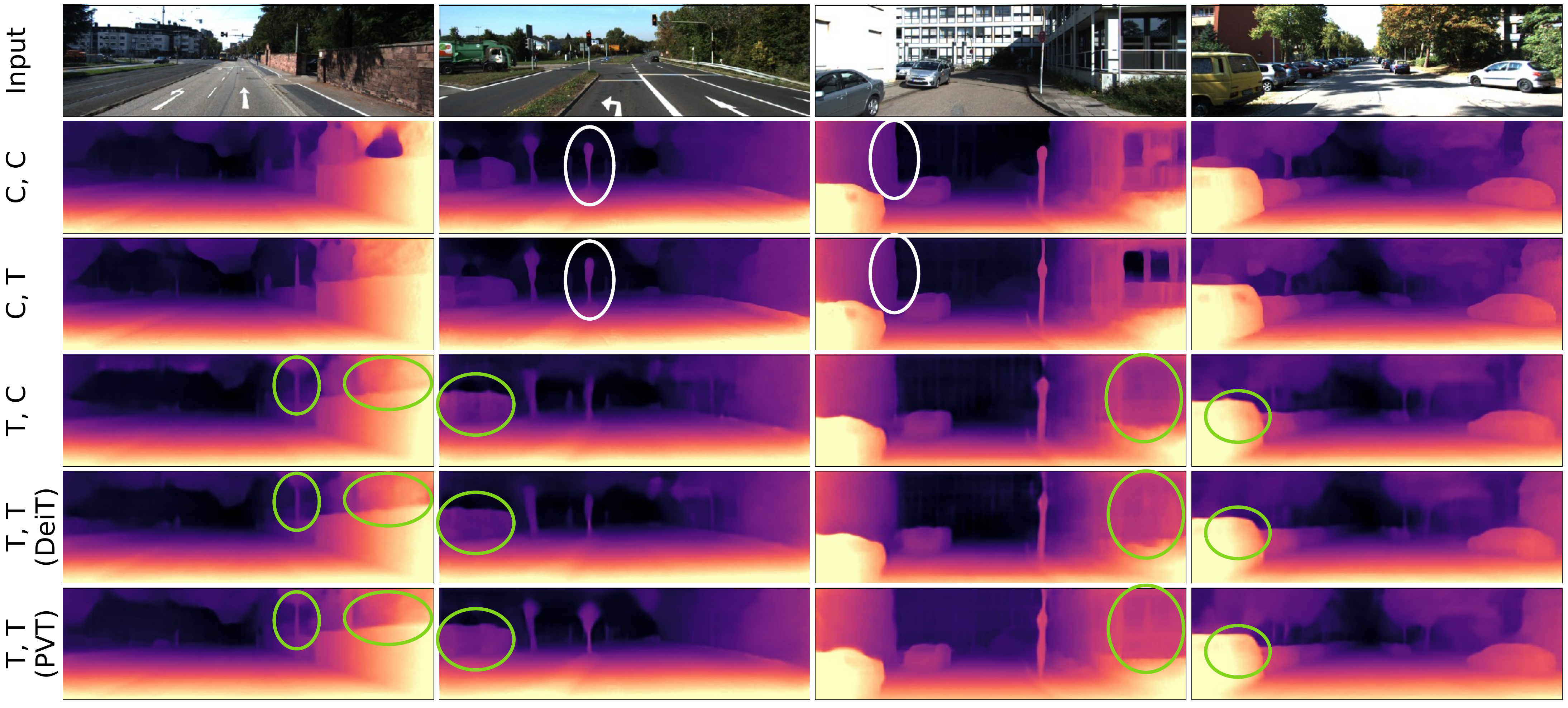}
  \caption{Disparity maps on KITTI Eigen for qualitative comparison of all four architecture combinations of depth and pose networks. Example regions where the global receptive field of transformers is advantageous are highlighted in green. Example areas where local receptive field of CNNs is advantageous are highlighted in white. Source:~\cite{visapp22}.
  } 
\label{fig:ablation_vis}
\end{figure}

Since unsupervised monocular depth estimation networks simultaneously train a pose network (see Equation \ref{eq:perspective_model}), we investigate the impact of each network's architecture on depth estimation. 
We consider CNN-based (C) and Transformer-based (T) networks for depth and pose estimation. The four resulting combinations of (Depth Network, Pose Network) architectures, in ascending order of impact of transformers on depth estimation, are (C, C), (C, T), (T, C), and (T, T). 
To compare the transformer-based architecture fairly with CNN-based networks, we utilize Monodepth2~\cite{godard2019digging} with ResNet-101~\cite{he2016deep} in the depth and pose encoders. All four combinations are trained thrice on the KITTI Eigen split using the settings described in Section \ref{subsec:details} and the known ground-truth camera intrinsics.

\subsubsection{On Performance.}
Table \ref{tab:ablation_table} shows the best results on depth estimation for each architecture combination of depth and pose networks.
We observe that MT-SfMLearner, i.e. the combination of transformer-based depth and pose networks, performs best under two of the \textit{error} metrics and two of the \textit{accuracy} metrics. The remaining combinations show comparable performance on all metrics.
Figure \ref{fig:ablation_vis} also shows more uniform estimates for larger objects, such as vehicles, vegetation, and buildings, when the depth is learned using transformers. Transformers also estimate depth more coherently for reflections from windows of vehicles and buildings. This is likely because of the larger receptive fields of the self-attention layers, which lead to more globally coherent predictions. On the other hand, convolutional networks produce sharper boundaries and perform better on thinner objects such as traffic signs and poles. This is likely due to the inherent inductive bias for spatial locality present in convolutional layers.

\subsubsection{On Robustness.}
We saw in the previous section that the different architecture combinations perform comparably on the independent and identically distributed (i.i.d) test set. However, networks that perform well on an i.i.d test set may still learn shortcut features that reduce robustness on out-of-distribution (o.o.d) datasets~\cite{geirhos2020shortcut}. 
Therefore, we study the robustness of each architecture combination. We report the mean RMSE across three training runs on the KITTI Eigen split test set for all experiments in this section.

\begin{figure}[t]
\centering
\includegraphics[width=\textwidth]{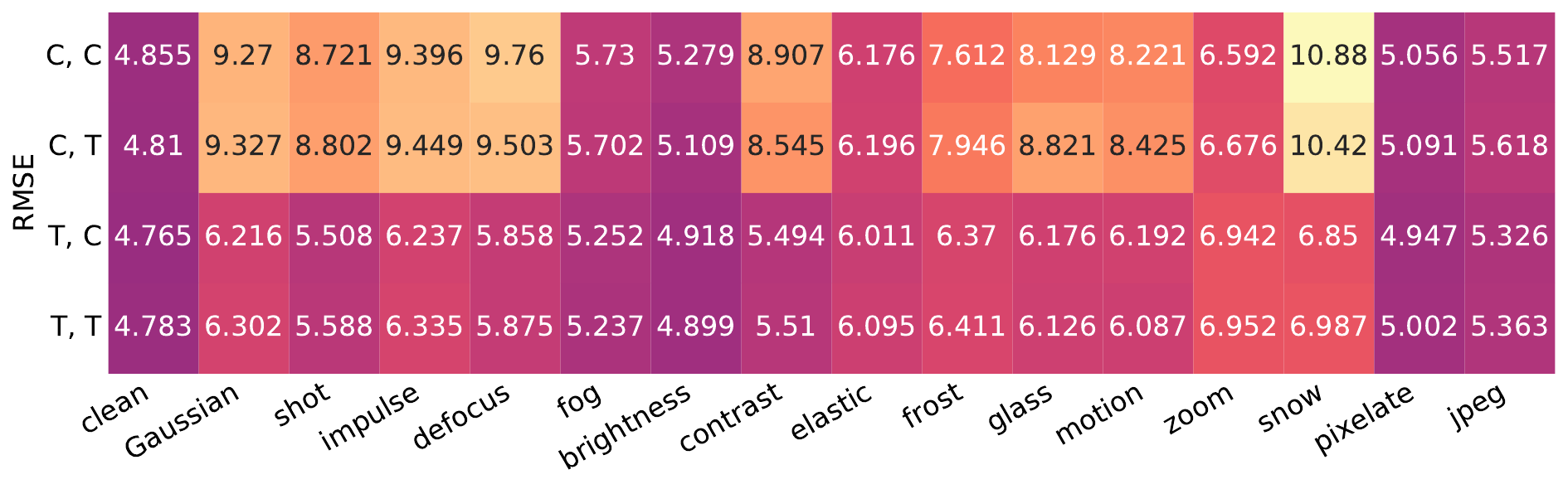}
\caption{RMSE for natural corruptions of KITTI Eigen test set for all four combinations of depth and pose networks. The i.i.d evaluation is denoted by \textit{clean}. Source:~\cite{visapp22}.
}
\label{fig:heatmap_corruption}
\end{figure}

\paragraph{Natural Corruptions:}
Following~\cite{hendrycks2019robustness} and ~\cite{michaelis2019dragon}, we generate 15 corrupted versions of the KITTI i.i.d test set at the highest severity($=5$). These corruptions are changes to images that correspond to variations expected in nature, such as those due to \textit{noise}  (Gaussian, shot, impulse), \textit{blur} (defocus, glass, motion, zoom), \textit{weather} (snow, frost, fog, brightness) and \textit{digital} (contrast, elastic, pixelate, JPEG).
We observe in Figure \ref{fig:heatmap_corruption} that learning depth with transformer-based architectures instead of CNN-based architecture leads to a significant improvement in the robustness to all natural corruptions.

\paragraph{Untargeted Adversarial Attack:}
\begin{figure}[ht]
\centering
\includegraphics[width=0.65\textwidth]{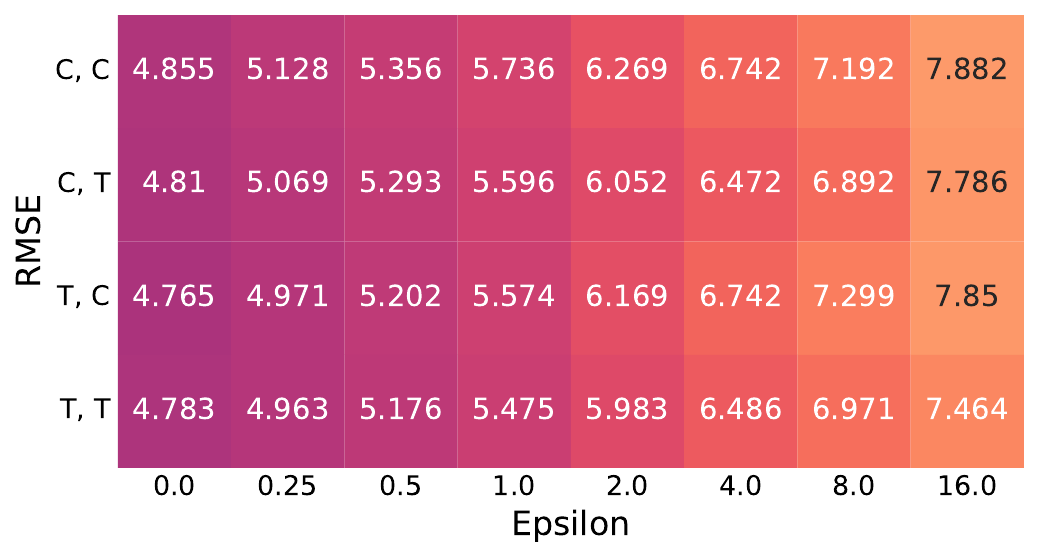}
\caption{RMSE for adversarial corruptions of KITTI Eigen test set generated using the PGD attack at all attack strengths ($0.0$ to $16.0$) for the four combinations of depth and pose networks. Attack strength $0.0$ refers to i.i.d evaluation. Source:~\cite{visapp22}.}
\label{fig:heatmap_pgd}
\end{figure}

Untargeted adversarial attacks make changes to input images that are imperceptible to humans to generate adversarial examples that can induce general prediction errors in neural networks. We employ Projected Gradient Descent (PGD)~\cite{DBLP:conf/iclr/MadryMSTV18} to generate untargeted adversarial examples from the test set at attack strength $\epsilon \in \{0.25, 0.5, 1.0, 2.0, 4.0, 8.0, 16.0\}$. The gradients are calculated with respect to the appearance-based training loss. Following ~\cite{kurakin2016adversarial}, the adversarial perturbation is computed over $min(\epsilon + 4, \lceil1.25\cdot\epsilon\rceil)$ iterations with a step size of $1$. When the test image is from the beginning or end of a sequence, the training loss is only calculated for the feasible pair of images. Figure \ref{fig:heatmap_pgd} demonstrates a general improvement in untargeted adversarial robustness when learning depth or pose with a transformer-based architecture instead of a CNN-based architecture.

\paragraph{Targeted Adversarial Attack:}

\begin{figure}[t]
\centering
\includegraphics[width=\textwidth]{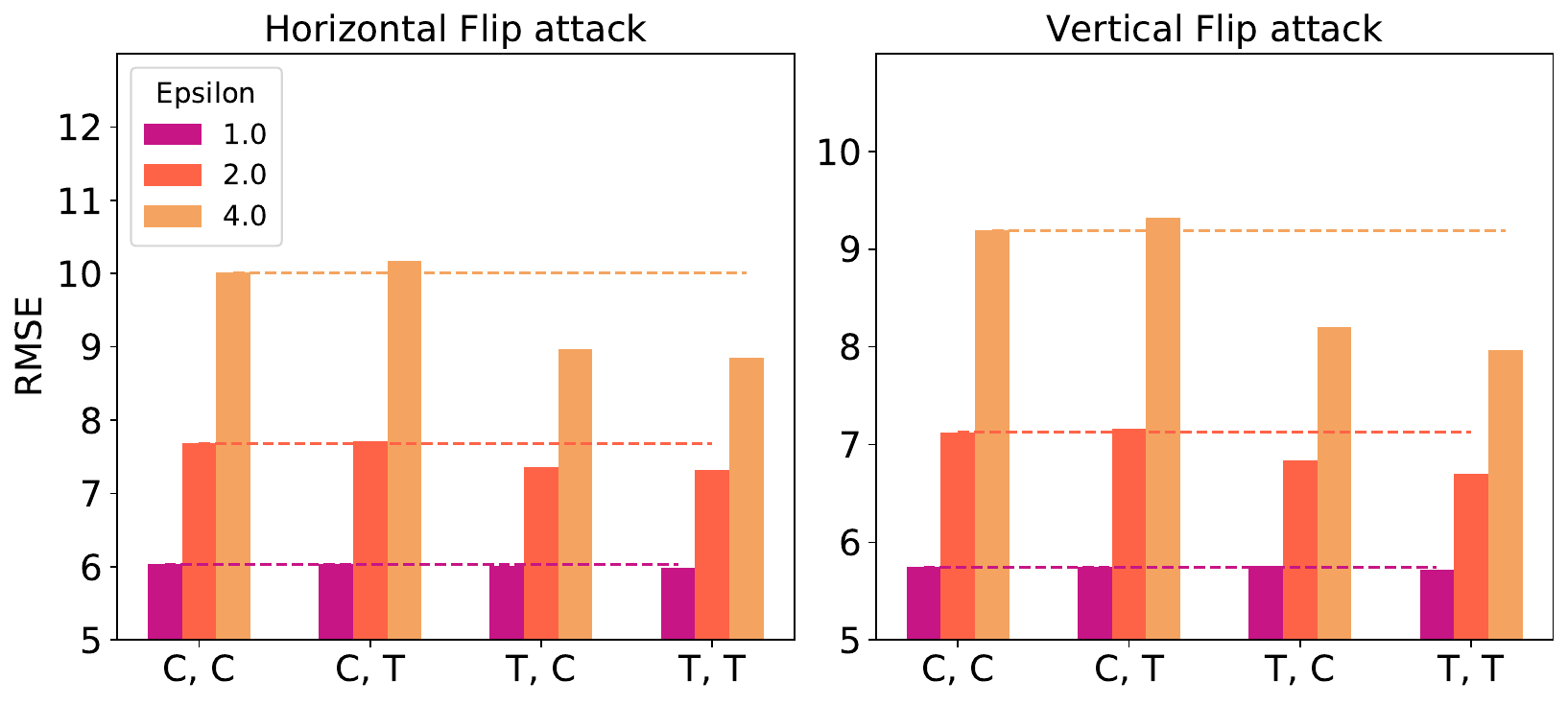}
\caption{RMSE for adversarial corruptions of KITTI Eigen test set generated using targeted horizontal and vertical flip attacks for all four combinations of depth and pose networks. Source:~\cite{visapp22}.
}
\label{fig:bar_flip}
\end{figure}
Finally, targeted adversarial attacks make changes imperceptible for humans to generate adversarial examples that can induce \textit{specific} prediction errors in neural networks. Deriving from ~\cite{wong2020targeted}, we generate targeted adversarial examples to fool the networks into predicting horizontally and vertically flipped estimates. To this end, we use the gradients with respect to the RMSE loss, where the targets are symmetrical horizontal and vertical flips of the predictions on the clean test set images. This evaluation is conducted at attack strength $\epsilon \in \{1.0, 2.0, 4.0\}$.
Figure \ref{fig:bar_flip} shows an improvement in robustness to targeted adversarial attacks when depth is learned using transformer-based architectures instead of CNN-based architectures. Furthermore, the combination where both depth and pose are learned using transformer-based architectures is the most robust.

Therefore, MT-SfMLearner, where depth and pose are learned with transformer-based architectures, provides the highest robustness against natural corruptions and untargeted and targeted adversarial attacks, according to studies on image classification~\cite{paul2021vision,bhojanapalli2021understanding}.
This can be attributed to their global receptive field, which allows for better adjustment to the localized 
deviations by accounting for the global context of the scene.

\subsection{Generalizability}
\label{subsec:generalizability}
The previous section showed that the use of transformers in the depth and pose estimation network contributes to the improved performance and robustness of depth estimation compared to their convolutional counterparts. We now examine whether this conclusion holds when examining on a different dataset or with a different transformer-based encoder. Note that the experiments herein directly show the comparison between (C,C) and (T,T). All comparisons assume known ground-truth camera intrinsics. 

\subsubsection{Different Dataset}
We compare on the DDAD dataset, which has scenes from a variety of locations, and has a ground truth depth of a longer range than KITTI. As earlier, we compare both the performance of depth estimation on i.i.d test set as well as the robustness to natural corruptions and adversarial attacks. 

\paragraph{Performance:} We report the best performance for each architecture on i.i.d in Table~\ref{tab:ddad_ablation_table}. Note that (T,T) outperforms (C,C) in almost all metrics. 
This confirms the generalizability of i.i.d performance of transformers across datasets. Additionally, we note that transformers also outperform the 3D convolutional PackNet architecture designed to handle long-range depth as curated within DDAD, despite a lack of inductive bias for the same in transformers.

\begin{table}[t]
\centering
\caption{Quantitative results on DDAD (complete). The best results are shown in bold.} 
\begin{tabular}{|c|c|c|c|c|c|c|c|}
\hline
\multirow{2}{*}{\textbf{Architecture}}  & \multicolumn{4}{c|}{\textbf{\cellcolor{red!25}Error$\downarrow$}} & \multicolumn{3}{c|}{\textbf{\cellcolor{blue!25}Accuracy$\uparrow$}} \\ \cline{2-8}
       & Abs Rel & Sq Rel & RMSE & RMSE log & $\delta<1.25$ & $\delta<1.25^2$ & $\delta<1.25^3$ \\ \hline \hline
     Packnet~\cite{guizilini20203d} & 0.178 & 7.521 & 14.605 & 0.254 & 0.831 & 0.928 & 0.963 \\ \hline \hline 
 C, C & \textbf{0.151}	& \textbf{3.346}	& 14.229 & 0.243 & 0.814 & 0.929 & 0.967 \\
  T,T & \textbf{0.151} & 3.821 & \textbf{14.162} & \textbf{0.237} & \textbf{0.820} & \textbf{0.935} & \textbf{0.970} \\
\hline
\end{tabular}
\label{tab:ddad_ablation_table}
\end{table}

\begin{figure}[t]
\centering
\includegraphics[width=\textwidth]{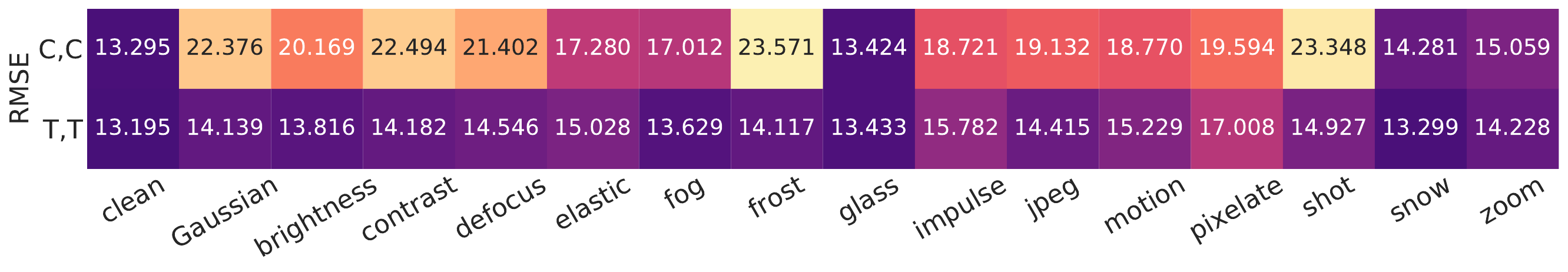}
\caption{RMSE for natural corruptions of DDAD. The i.i.d evaluation is denoted by \textit{clean}.}
\label{fig:ddad_heatmap_corruption}
\end{figure}

\begin{figure}[t]
\centering
\includegraphics[width=0.8\textwidth]{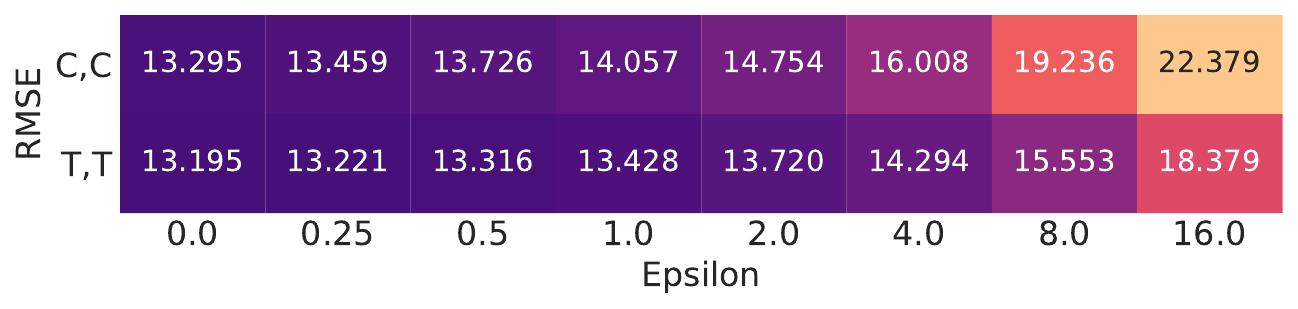}
\caption{RMSE for untargeted attacks on DDAD. The i.i.d evaluation is denoted by \textit{clean}.}
\label{fig:ddad_heatmap_pgd}
\end{figure}

\begin{figure}[b]
\centering
\includegraphics[width=0.75\textwidth]{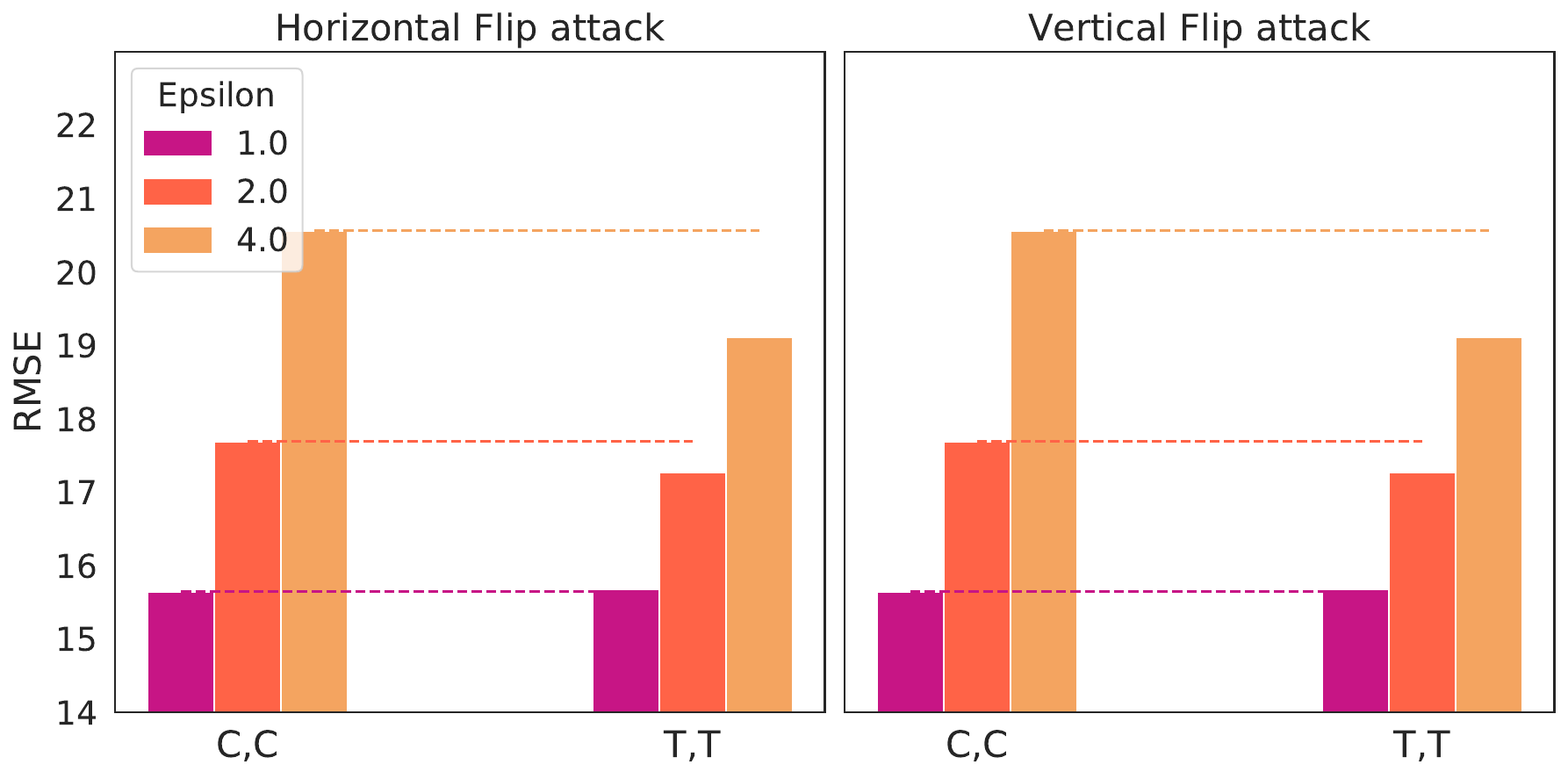}
\caption{RMSE for adversarial corruptions of DDAD generated using horizontal and vertical flip attacks. The i.i.d evaluation is denoted by \textit{clean}.}
\label{fig:ddad_adv}
\end{figure}

\paragraph{Robustness:}

Next, we compare the robustness of the architectures against natural corruptions and adversarial attacks on the DDAD dataset. Note that this analysis is performed on a subset of the DDAD test set, consisting of a randomly selected subset of $11$ of the $50$ test sequences. In particular, these sequences are \#\{151, 154, 156, 167, 174, 177, 179, 184, 192, 194, 195\}. For robustness evaluation, we report the mean RMSE across three training runs. 

Figure~\ref{fig:ddad_heatmap_corruption} compares the robustness of the architectures to natural corruptions. We find that transformers are significantly better at handling 
across natural corruptions.
Figure~\ref{fig:ddad_heatmap_pgd} further compares the robustness of architectures with untargeted adversarial perturbations. We find that transformers are also better against untargeted adversarial attacks, with the difference becoming more pronounced as the attack strength increases. Finally, Figure~\ref{fig:ddad_adv} compares the robustness of the architectures with the targeted horizontal and vertical flip attacks. Again, we find that transformers are also more robust to targeted attacks. 

The above experiments confirm the generalizability of robustness of transformers across datasets.

\subsubsection{Different Encoder Backbone}
While we utilized the Data Efficient Image Transformer (DeiT) backbone for the encoder in previous experiments, we now evaluate MT-SfMLearner with the Pyramid Vision Transformer (PVT) backbone. As earlier, we compare both the performance of depth estimation on i.i.d test set as well as the robustness to natural corruptions and adversarial attacks for the KITTI Eigen split. 

\paragraph{Performance:} We report the best performance for each architecture on i.i.d in Table~\ref{tab:pvt_ablation_table}. Note that (T,T) with PVT outperforms not only (C,C) but also (T,T) with DeiT in all metrics. This can be attributed to the design of PVT, built particularly for dense prediction tasks, such as depth estimation. 
Figure~\ref{fig:ablation_vis} further shows continued globally coherent predictions for large objects and reflections for PVT. Furthermore, PVT improves on thin structures over DeiT, likely due to its overlapping patch embedding, which allows for better learning of local information. 
This confirms the generalizability of i.i.d performance of transformers across different encoder backbones.  

\begin{table}[t]
\centering
\caption{Quantitative results on KITTI Eigen with modified encoder backbone. The best results are shown in bold.} 
\begin{tabular}{|c|c|c|c|c|c|c|c|}
\hline
\multirow{2}{*}{\textbf{Architecture}}  & \multicolumn{4}{c|}{\textbf{\cellcolor{red!25}Error$\downarrow$}} & \multicolumn{3}{c|}{\textbf{\cellcolor{blue!25}Accuracy$\uparrow$}} \\ \cline{2-8}
       & Abs Rel & Sq Rel & RMSE & RMSE log & $\delta<1.25$ & $\delta<1.25^2$ & $\delta<1.25^3$ \\ \hline \hline
 C, C &  0.111 & 0.897 & 4.865 & 0.193 & 0.881 & 0.959 & 0.980 \\
 T,T (DeiT) &  0.112 & 0.838 & 4.771 & 0.188 & 0.879 & 0.960 & \textbf{0.982} \\
  T,T (PVT) & \textbf{0.107}   &  \textbf{0.780}   &  \textbf{4.537}   &  \textbf{0.183}   &  \textbf{0.890}   &  \textbf{0.963}   &  \textbf{0.982} \\
  \hline
\end{tabular}
\label{tab:pvt_ablation_table}
\end{table}

\paragraph{Robustness:} Next, we compare the robustness of the architectures against natural corruptions and adversarial attacks. For robustness evaluation, we report the mean RMSE across three training runs. 

\begin{figure}[b]
\centering
\includegraphics[width=\textwidth]{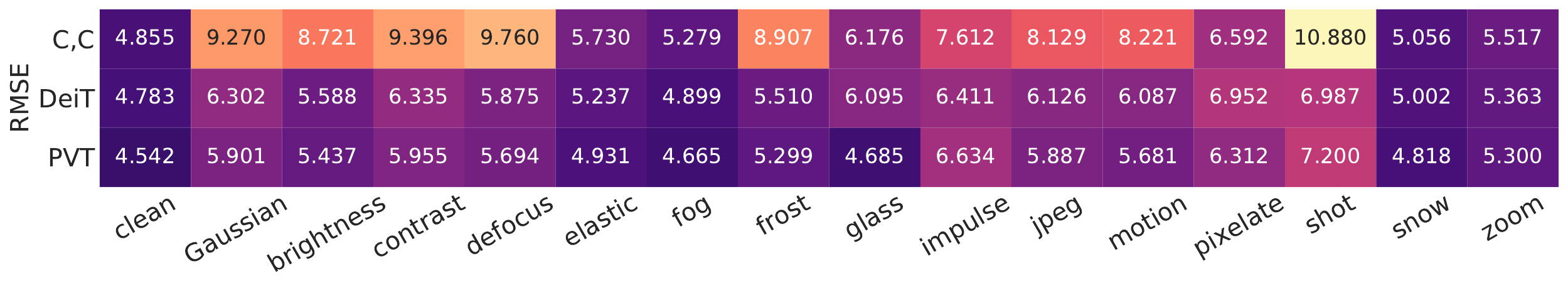}
\caption{RMSE for natural corruptions of KITTI Eigen including PVT. The i.i.d evaluation is denoted by \textit{clean}.}
\label{fig:pvt_heatmap_corruption}
\end{figure}

\begin{figure}[ht]
\centering
\includegraphics[width=0.75\textwidth]{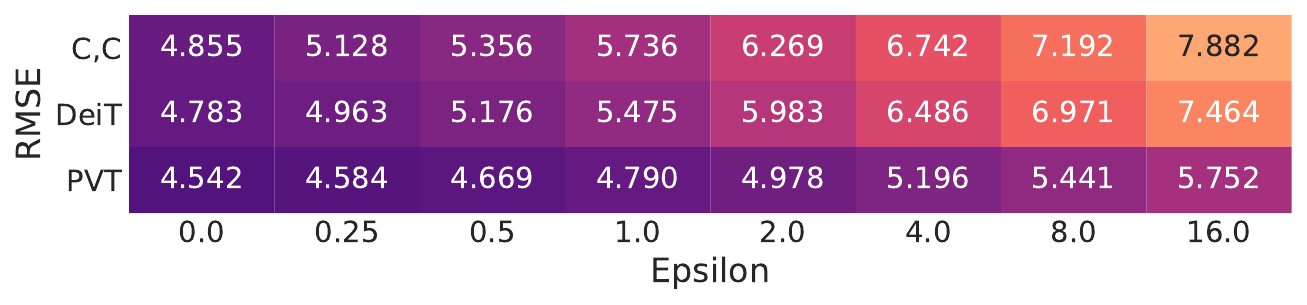}
\caption{RMSE for untargeted attacks on KITTI Eigen, including T, T (PVT). The i.i.d evaluation is denoted by \textit{clean}.
}
\label{fig:pvt_heatmap_pgd}
\end{figure}

\begin{figure}[t]
\centering
\includegraphics[width=0.85\textwidth]{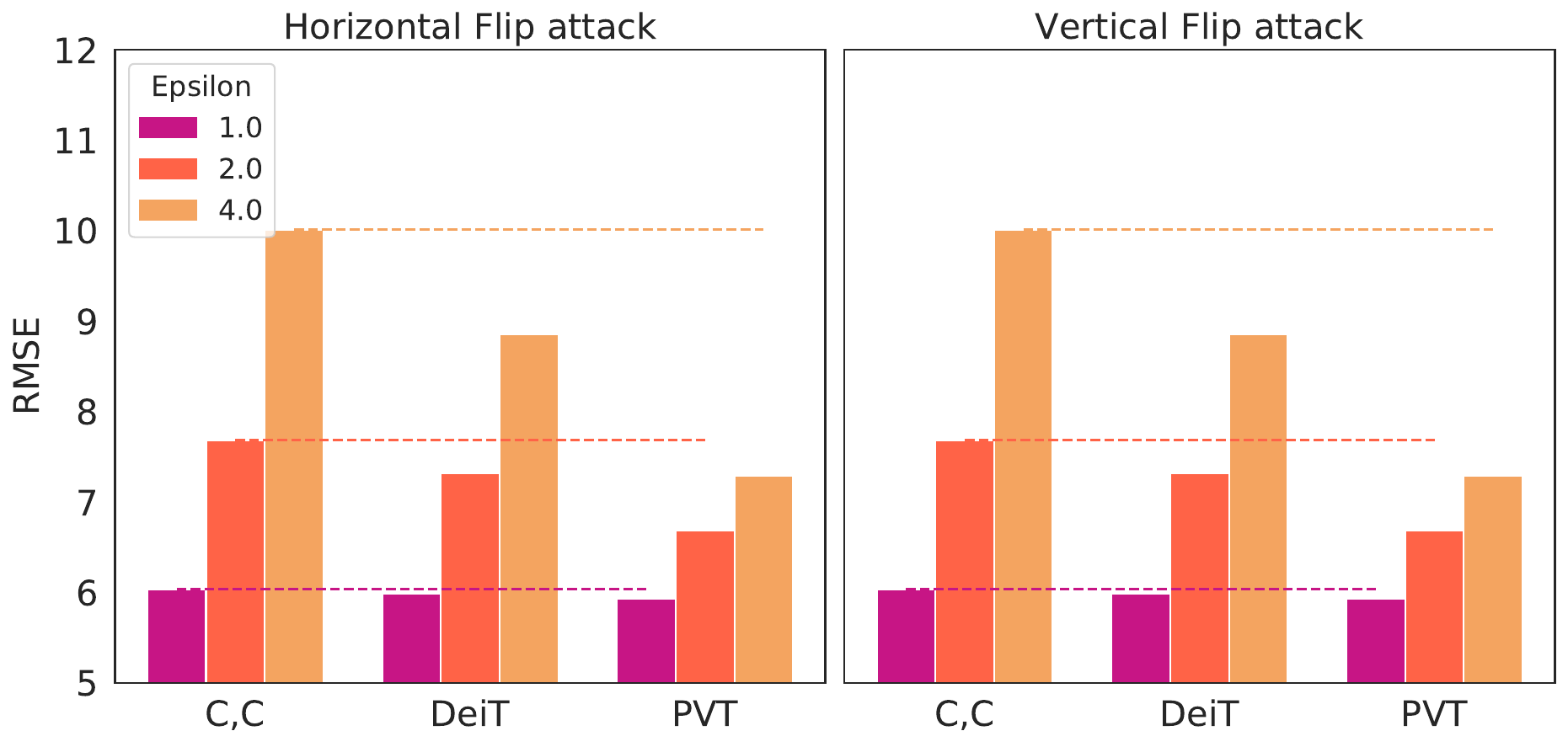}
\caption{RMSE for adversarial corruptions of KITTI Eigen including T, T (PVT) generated using horizontal and vertical flip attacks. The i.i.d evaluation is denoted by \textit{clean}.}
\label{fig:pvt_adv}
\end{figure}

Figures~\ref{fig:pvt_heatmap_corruption}, ~\ref{fig:pvt_heatmap_pgd}, and ~\ref{fig:pvt_adv} compare the robustness of the architectures to natural corruptions, untargeted adversarial attack,  and targeted horizontal and vertical flip attack. We find that both transformer architectures are significantly better at handling natural corruptions as well as untargeted and targeted adversarial attacks. 
We hypothesize that PVT has much higher robustness than even DeiT due to its spatial feature pyramid and overlapping patch embedding that helps to maintain local continuity in the image.

Therefore, the above experiments also confirm the generalizability of the robustness of transformers across encoder backbones.

\subsection{Auxiliary Tasks}
\label{sec:auxiliary}

\begin{table}[b]
\centering
\caption{Percentage error for intrinsics prediction. Source:~\cite{visapp22}
}
\begin{tabular}{|c|cccc|}
\hline
\multirow{2}{*}{\textbf{Network}} & \multicolumn{4}{c|}{\textbf{\cellcolor{red!25} Error(\%) $\downarrow$}} \\ 
 \cline{2-5} & $f_{x}$ & $c_{x}$ & $f_{y}$ & $c_{y}$ \\ \hline \hline
  C,C  & -1.889 & -2.332 & 2.400 & -9.372  \\
  \hline
  T,T  & -1.943 & -0.444 & 3.613 & -16.204 \\
  \hline
\end{tabular}
\label{tab:intrinsics_table}
\end{table}

Unsupervised monocular SfM requires access to relative pose between image pairs and camera intrinsics corresponding to the input, in addition to depth. As discussed in Section~\ref{sec:method}, a network is simultaneously trained for the pose and (optionally) camera intrinsics estimation along with the depth estimation network. 
While we have studied depth estimation in detail in the previous subsections, here we examine if the improved performance with transformers comes at the expense of the performance on auxiliary tasks.

\paragraph{Intrinsics Estimation:}
In Table~\ref{tab:intrinsics_table}, we examine the accuracy of our proposed intrinsics estimation network on the KITTI Eigen split. We observe that both the CNN-based and transformer-based architectures result in a low percentage error on the focal length and principal point. The performance and robustness of depth estimation with learned camera intrinsics are discussed in Section~\ref{sec:depth_with_learned_K}.

\begin{figure}[ht]
\centering
\begin{tabular}{cc}
     \includegraphics[width=.5\textwidth]{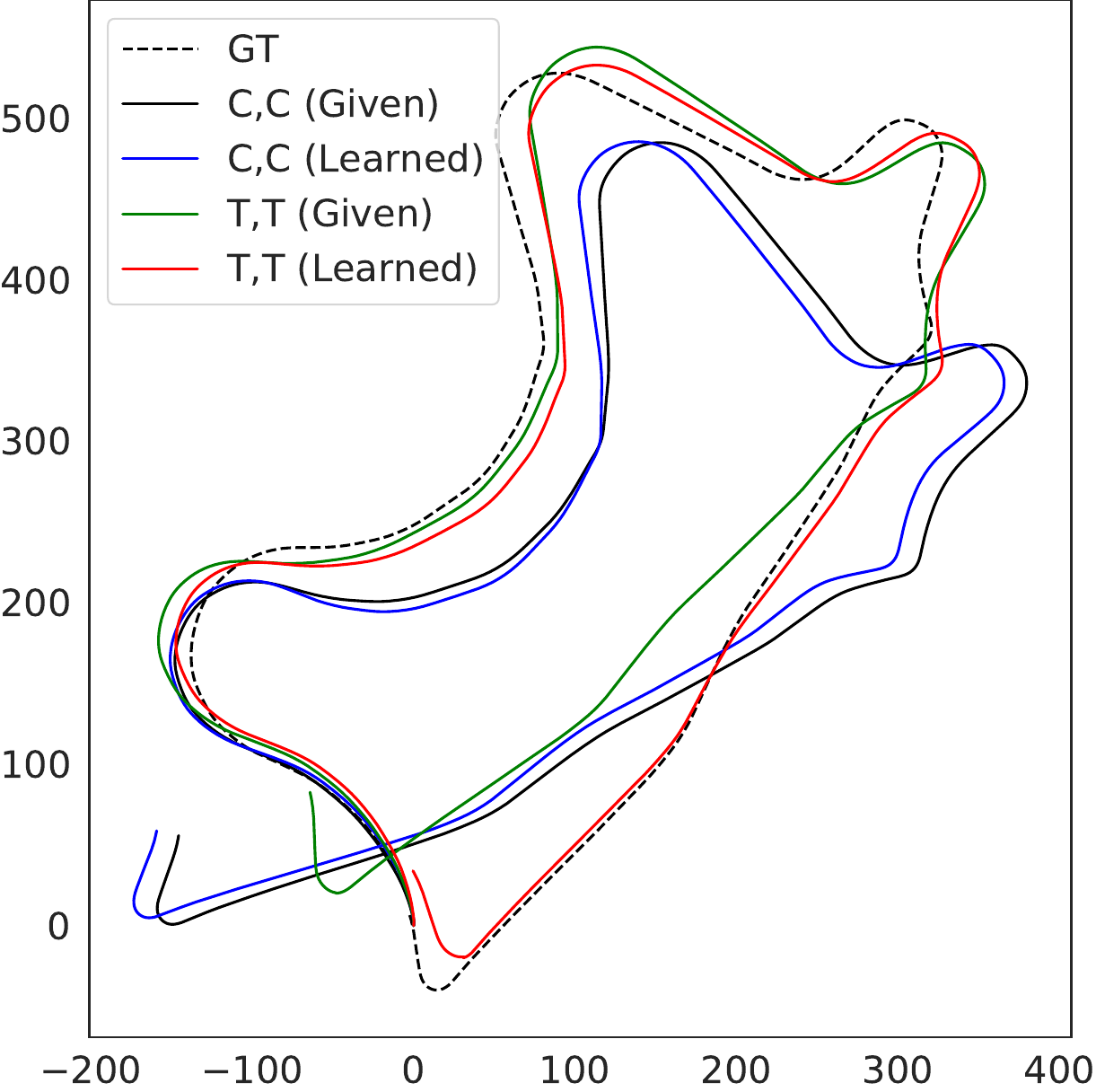} \\  \includegraphics[width=.5\textwidth]{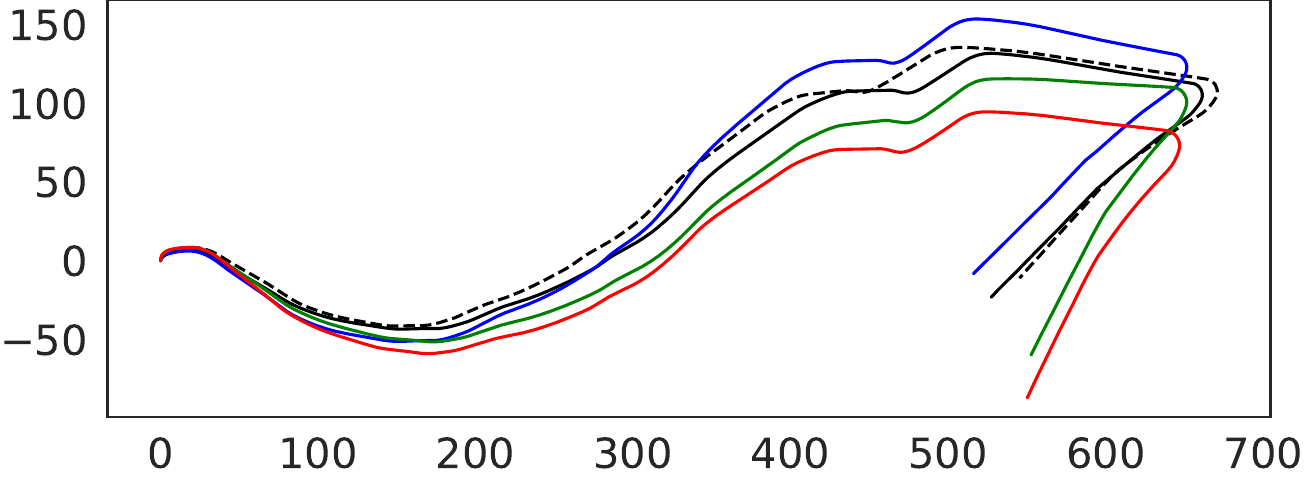} \\
\end{tabular}
\caption{Origin-aligned trajectories on KITTI Odom split for (C,C) and (T,T) showing the impact of learning intrinsics.
}
\label{fig:trajectories}
\end{figure}

\paragraph{Pose Estimation:}
 In Table~\ref{tab:intrinsics_table_odom}, we examine the accuracy of the pose estimation network on the KITTI Odom split, including when the intrinsics are unknown. We observe that both the translation and rotation errors for sequence 09 are lower with (T,T) than with (C,C) when the camera intrinsics are given. However, the opposite is true for Sequence 10. We also observe that the translation and rotation errors for both sequences are similar to when the ground truth intrinsics are known a priori. Figure~\ref{fig:trajectories} visualizes the predicted trajectories with both architectures, including when the intrinsics are unknown.

\begin{table}[t]
\centering
\caption{Impact of estimating intrinsics on pose estimation for the KITTI Odom split.}
\begin{tabular}{|c|c|cc|cc|}
\hline
\multirow{2}{*}{\textbf{Network}} & \multirow{2}{*}{\textbf{Intrinsics}}  &
\multicolumn{2}{c|}{\textbf{Seq 09}} & \multicolumn{2}{c|}{\textbf{Seq 10}} \\  \cline{3-6} \cline{3-6} 
& & \cellcolor{red!25} $t_{err} (\%) \downarrow$ & \cellcolor{red!25} $r_{err} (^{\circ}/100m)\downarrow$ & \cellcolor{red!25} $t_{err}(\%) \downarrow$ & \cellcolor{red!25} $r_{err} (^{\circ}/100m) \downarrow$ \\ \hline \hline
  \multirow{2}{*}{C,C} & Given & 10.376 & 3.237 & 7.784  & 2.444  \\
    & Learned  & 14.052 & 4.047 & 11.780 & 3.338 \\
  \hline 
  \multirow{2}{*}{T,T} & Given & 6.998 & 2.181 & 8.983 & 3.666 \\
    & Learned & 7.624 & 2.099 & 9.537 & 3.962 \\
  \hline
\end{tabular}
\label{tab:intrinsics_table_odom}
\end{table}

Overall, we conclude that the benefits of using transformers do not come at the expense of its performance on intrinsics or pose estimation. Note that while auxiliary tasks perform well, traditional methods continue to dominate with several methods combining deep learning approaches with the traditional methods~\cite{bian2021unsupervised,chawla2020crowdsourced}

\subsection{Depth Estimation with Learned Camera Intrinsics}
\label{sec:depth_with_learned_K}

We analyze the impact on performance and robustness of depth estimation when the camera intrinsics are unknown a priori and the network is trained to estimate it. The experiments are performed on the KITTI Eigen split.

\begin{figure}[t]
\centering
 \includegraphics[width=\textwidth]{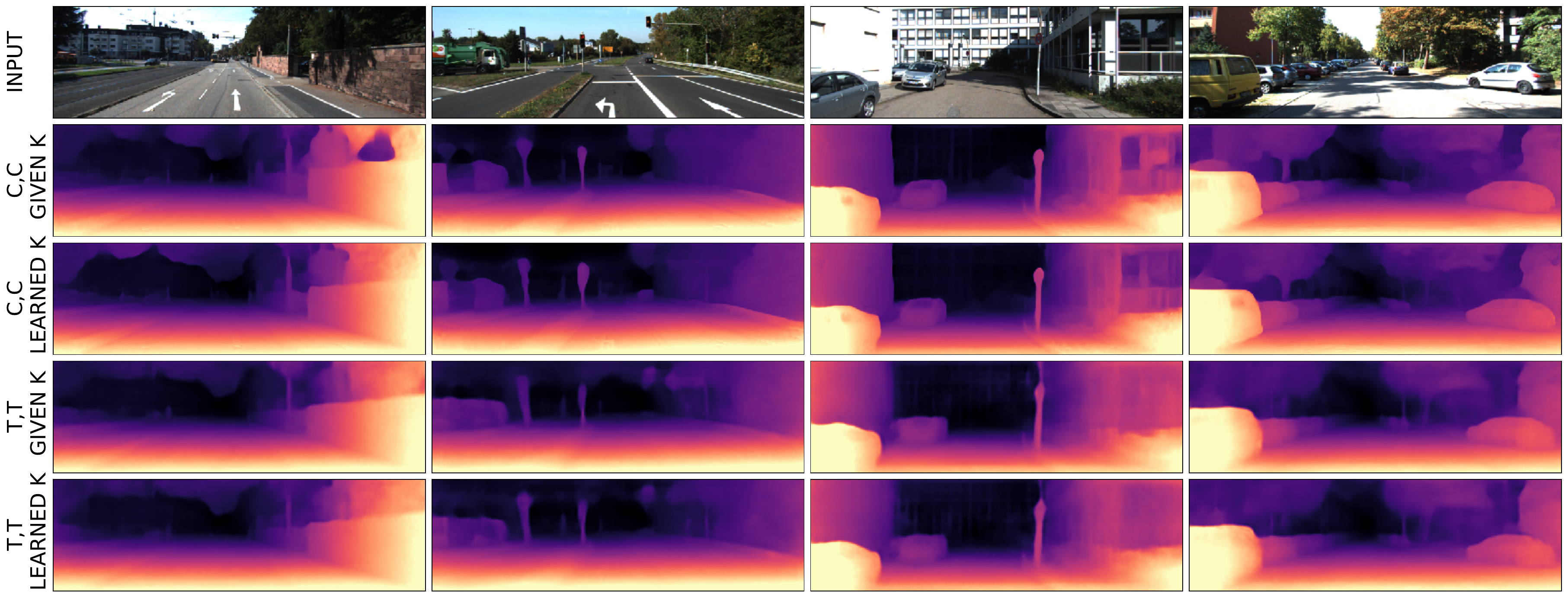}
  \caption{Disparity maps for qualitative comparison on KITTI, when trained with and without intrinsics (K). The second and fourth rows are same as the second and the fifth rows in Figure \ref{fig:ablation_vis}. Source:~\cite{visapp22}.} 
\label{fig:intrinsics_depth_vis}
\end{figure}

\paragraph{Performance:} 
Table~\ref{tab:intrinsics_depth} compares the accuracy and error for depth estimation when intrinsics are learned and when they are given a priori. We observe that depth error and accuracy for transformer-based architectures continue to be better than those for CNN-based architectures, even when the intrinsics are learned. Additionally, we observe that the depth error and accuracy metrics are both similar to those when the ground truth intrinsics are known. This is also substantiated in Figure~\ref{fig:intrinsics_depth_vis} where learning of the intrinsics does not cause artifacts in depth estimation.

\begin{table}[bt]
\centering
\caption{Impact of estimating intrinsics on depth estimation for KITTI Eigen split. Source:~\cite{visapp22}. 
} 
\begin{tabular}{|c|c|cccc|ccc|}
\hline
\multirow{2}{*}{\textbf{Network}} & \multirow{2}{*}{\textbf{Intrinsics}}  &
\multicolumn{4}{c|}{\textbf{\cellcolor{red!25} Depth Error$\downarrow$}} &
\multicolumn{3}{c|}{\textbf{\cellcolor{blue!25}Depth Accuracy$\uparrow$}} \\ 
\cline{3-9} & & Abs Rel & Sq Rel & RMSE & RMSE log & $\delta<1.25$ & $\delta<1.25^2$ & $\delta<1.25^3$ \\ \hline \hline
\multirow{2}{*}{}
  \multirow{2}{*}{C,C} & Given & 0.111 & 0.897 & 4.865 & 0.193 & 0.881 & 0.959 & 0.980 \\
    & Learned  & 0.113 & 0.881 & 4.829 & 0.193 & 0.879 & 0.960 & 0.981 \\
  \hline
  \multirow{2}{*}{T,T} & Given &  0.112 & 0.838 & 4.771 & 0.188 & 0.879 & 0.960 & 0.982 \\
    & Learned & 0.112 & 0.809 & 4.734 & 0.188 & 0.878 & 0.960 & 0.982\\
  \hline
\end{tabular}
\label{tab:intrinsics_depth}
\end{table}

\begin{figure}[hbt]
\centering
  \includegraphics[width=0.6\linewidth]{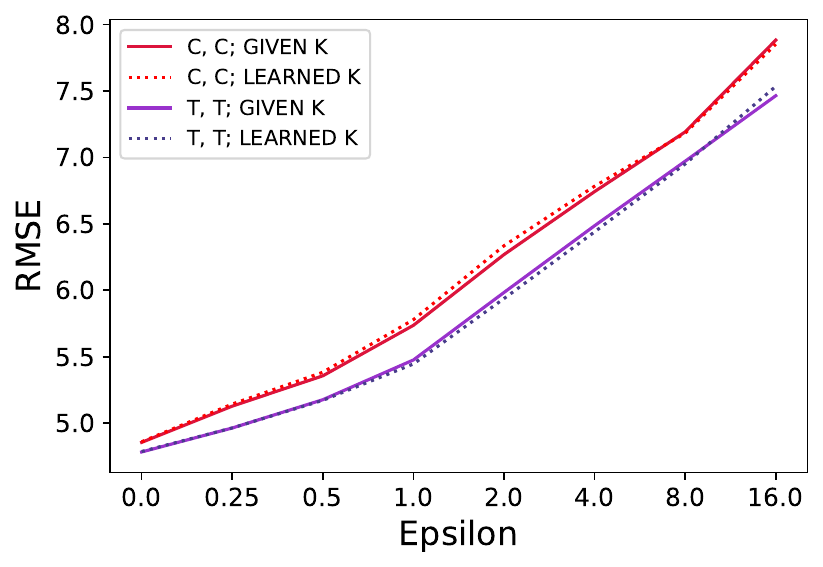}
  \caption{RMSE for adversarial corruptions of KITTI generated using untargeted PGD attack, when trained with and without ground truth intrinsics (K). Source:~\cite{visapp22}.} 
\label{fig:adversarial_intrinsics}
\end{figure}

 \paragraph{Robustness:} As before, we also evaluate networks trained with learned intrinsics for their robustness against natural corruptions, untargeted attack, and targeted adversarial attacks. We report the mean RMSE ($\mu$RMSE) across all corruptions and for all attacks strengths in targeted adversarial attacks in Table~\ref{tab:intrinsics_corr}, the RMSE on the untargeted adversarial attack in Figure~\ref{fig:adversarial_intrinsics} averaged over three runs. We observe that both the architectures maintain similar robustness against natural corruptions and adversarial attacks when the intrinsics are learned simultaneously as opposed to when intrinsics are known a priori. Additionally, similar to the scenario with known ground truth intrinsics, transformers with learned intrinsics is more robust than its convolutional counterpart.  
 
 \begin{table}[t]
\centering
\caption{Mean RMSE ($\mu$RMSE) for natural corruptions, horizontal (H) and vertical (V) adversarial flips of KITTI, when trained with and without ground truth intrinsics. Source:~\cite{visapp22}.}
\begin{tabular}{|c|c|c|cc|}
\hline
& &  Natural corruptions &  \multicolumn{2}{c|}{Adversarial attack}\\
\textbf{Architecture}  & \textbf{Intrinsics} & \textbf{\cellcolor{red!25}$\mu$RMSE$\downarrow$} & \textbf{\cellcolor{red!25}$\mu$RMSE$\downarrow$ (H)} & \textbf{\cellcolor{red!25}$\mu$RMSE$\downarrow$ (V)}    \\ \hline \hline
\multirow{2}{*}{C, C}   & Given      & 7.683  & 7.909 & 7.354 \\
                        & Learned    & 7.714  & 7.641 & 7.196 \\ \hline
\multirow{2}{*}{T, T}   & Given      & 5.918  & 7.386 & 6.795 \\
                        & Learned    & 5.939  & 7.491 & 6.929 \\ \hline
\end{tabular}
\label{tab:intrinsics_corr}
\end{table}

\subsection{Efficiency}
In order to deploy the architectures for use in robots and autonomous driving systems, it is important to examine their suitability for real-time application. Thus, we evaluate the networks on their computational and energy efficiency.

\begin{table}[htbp]
\centering
\caption{Inference Speed and Energy Consumption for depth, pose, and intrinsics estimation using CNN- and transformer-based architectures. Source:~\cite{visapp22}.}
\begin{tabular}{|c|c|c|c|}
\hline 
\textbf{Architecture}  & \textbf{Estimate} &  \textbf{\cellcolor{blue!25}Speed$\uparrow$} & \textbf{\cellcolor{red!25}Energy$\downarrow$}\\ \hline \hline
\multirow{2}{*}{C,C}  & Depth  & 84.132      & 3.206 \\ 
 & Intrinsics/ Pose &  97.498      & 2.908 \\ \hline
\multirow{2}{*}{T,T} & Depth   &  40.215      & 5.999 \\ 
 & Intrinsics/ Pose & 60.190      & 4.021 \\ \hline
\end{tabular}
\label{tab:efficiency}
\end{table}

Table~\ref{tab:efficiency} reports the average speed and the average energy consumption during inference for depth, as well as pose and intrinsics networks for both architectures. These metrics are computed over 10,000 forward passes on NVidia GeForce RTX 2080 Ti. We observe that both architectures run in real-time with an inference speed $>$ 30 fps. Nevertheless, the energy consumption and computational costs for transformer-based architecture are higher than those of its CNN-based counterpart.

\subsection{Comparing Performance}
\label{subsec:depth_table}
Having established the benefits of transformers for depth estimation, we now evaluate MT-SfMLearner, where both depth and pose networks are transformer-based, with contemporary neural networks for their error and accuracy on unsupervised monocular depth estimation. Note that we do not compare with methods that use ground-truth depth or semantic labels during training. We also do not compare against methods that use multiple frames for depth estimation. In Table~\ref{tab:depth_table}, we observe that MT-SfMLearner (DeiT) achieves comparable performance against other methods including those with a heavy encoder such as ResNet-101~\cite{johnston2020self} and PackNet with 3D convolutions~\cite{guizilini20203d}. We also observe that MT-SfMLearner (PVT) outperforms these methods, including contemporary transformer-based methods such as MonoFormer~\cite{bae2022monoformer}. 

\begin{table}[htbp]
\centering
\caption{Quantitative results comparing MT-SfMLearner with existing methods on KITTI Eigen split. For each category of image sizes, the best results are displayed in bold, and the second best results are underlined. Table modified from ~\cite{visapp22}.}
\resizebox{\textwidth}{!}{
\begin{tabular}{|l|c|c|c|c|c|c|c|c|}
\hline
\multirow{2}{*}{\textbf{Methods}} & \multirow{2}{*}{\textbf{Resolution}} & \multicolumn{4}{c|}{\textbf{\cellcolor{red!25}Error$\downarrow$}} & \multicolumn{3}{c|}{\textbf{\cellcolor{blue!25}Accuracy$\uparrow$}} \\ \cline{4-9}
     &  & Abs Rel & Sq Rel & RMSE & RMSE log & $\delta<1.25$ & $\delta<1.25^2$ & $\delta<1.25^3$ \\ \hline \hline
  
   CC~\cite{ranjan2019competitive} & 832$\times$256 & 0.140 & 1.070 & 5.326 & 0.217 & 0.826 & 0.941 & 0.975 \\
   SC-SfMLearner~\cite{bian2019unsupervised} & 832$\times$256 & 0.137 & 1.089 & 5.439 & 0.217 & 0.830 & 0.942 & 0.975 \\
   Monodepth2 
  ~\cite{godard2019digging} & 640$\times$192 & 0.115 & 0.903 & 4.863 & 0.193 & 0.877 & 0.959 & 0.981 \\
   SG Depth~\cite{klingner2020selfsupervised} & 640$\times$192 & 0.117 & 0.907 & 4.844 & 0.194 & 0.875 & 0.958 & 0.980 \\
   PackNet-SfM~\cite{guizilini20203d} & 640$\times$192 & 0.111 & 0.829 & 4.788 & 0.199 & 0.864 & 0.954 & 0.980 \\
   Poggi et. al~\cite{poggi2020uncertainty} & 640$\times$192 & 0.111 & 0.863 & 4.756 & 0.188 & 0.881 & 0.961 & \underline{0.982} \\
   Johnston \& Carneiro~\cite{johnston2020self} & 640$\times$192 & \textbf{0.106} & 0.861 & 4.699 &  0.185 & \underline{0.889} & \underline{0.962} & \underline{0.982} \\
   HR-Depth~\cite{lyu2020hr} & 640$\times$192 & 0.109 & \underline{0.792} & 4.632 & 0.185 & 0.884 & \underline{0.962} & \textbf{0.983} \\  
   G2S~\cite{chawlavarma2021multimodal} & 640$\times$192 & 0.112 & 0.894 & 4.852 & 0.192 & 0.877 & 0.958 & 0.981 \\
   MonoFormer~\cite{bae2022monoformer} & 640$\times$192 & 0.108 & 0.806 & \underline{4.594} & \underline{0.184} & 0.884 & \textbf{0.963} & \textbf{0.983} \\ \hline
   \textbf{MT-SfMLearner (DeiT)} & 640$\times$192 & 0.112 & 0.838 & 4.771 & 0.188 & 0.879 & 0.960 & \underline{0.982} \\
   \textbf{MT-SfMLearner (PVT)} & 640$\times$192 & \underline{0.107} & \textbf{0.780} & \textbf{4.537} & \textbf{0.183} & \textbf{0.890} & \textbf{0.963} & \underline{0.982} \\
  \hline
\end{tabular}}
\label{tab:depth_table}
\end{table}
\section{Conclusion}

This work investigates the impact of transformer-based architecture on the unsupervised monocular Structure-from-Motion (SfM). 
We demonstrate that learning both depth and pose using transformer-based architectures leads to highest performance and robustness in depth estimation across multiple datasets and transformer encoders. We additionally establish that this improvement in depth estimation doesn't come at the expense of auxiliary tasks of pose and intrinsics estimation.
We also show that transformer-based architectures predict uniform and coherent depths, especially for larger objects, whereas CNN-based architectures provide local spatial-bias, especially for thinner objects and around boundaries. Moreover, our proposed intrinsics estimation module predicts intrinsics with low prediction error while maintaining performance and robustness on depth estimation. However, transformer-based architectures are more computationally demanding and have lower energy efficiency than their CNN-based counterpart.
Thus, we contend that this work assists in evaluating the trade-off between performance, robustness, and efficiency of unsupervised monocular SfM for selecting the suitable architecture.

\bibliographystyle{splncs04}
\bibliography{ref}

\end{document}